\documentclass[11pt,a4paper]{article}
\usepackage[a4paper, total={6in, 10in}]{geometry}
\usepackage{times}
\usepackage{latexsym}
\usepackage{amsmath}
\usepackage[ruled,vlined]{algorithm2e}
\usepackage{cleveref}
\usepackage{graphicx}
\usepackage{booktabs}
\usepackage{subcaption}
\usepackage{natbib}

\title{Large-Scale Multi-Document Summarization with Information Extraction and Compression}

\author{Ning Wang \\
  Northwestern University \\
  \texttt{ningwang2023@u.northwestern.edu} \\
  Han Liu \\
  Northwestern University \\
  \texttt{hanliu@northwestern.edu} \\
  Diego Klabjan \\
  Northwestern University \\
  \texttt{dklabjan@northwestern.edu} \\
  }

\date{}

\begin{document}
\maketitle
\begin{abstract}
We develop an abstractive summarization framework independent of labeled data for multiple heterogeneous documents.
Unlike existing multi-document summarization methods, our framework processes documents telling different stories instead of documents on the same topic. 
We also enhance an existing sentence fusion method with a uni-directional language model to prioritize fused sentences with higher sentence probability with the goal of increasing readability.
Lastly, we construct a total of twelve dataset variations based on CNN/Daily Mail and the NewsRoom datasets, where each document group contains a large and diverse collection of documents to evaluate the performance of our model in comparison with other baseline systems.
Our experiments demonstrate that our framework outperforms current state-of-the-art methods in this more generic setting.

\end{abstract}

\section{Introduction}
The task of text summarization focuses on extracting the most salient information in one or a collection of documents and formulate them in a concise and readable way. 
While most of the existing works on multi-document summarization focus on \textit{homogeneous} document groups with information overlap, a more generic and also practical case is to have \textit{heterogeneous} document groups with more independent documents and only a minor information overlap.
For example, in contrast to traditional multi-document methods capable of summarizing multiple reviews for a particular product, one use case of such a generic approach is to summarize a group of news articles in finance.
Another case of heterogeneous multi-document summarization is summarizing research articles in a specific field, such as work studying COVID-19. 
In both cases, articles contain different stories or arguments with the possibility of overlapping information.

The nature of heterogeneous document groups necessitates different methodologies. 
If articles in a group discuss the same topic, then such overlapping information could be utilized to determine the content of the summary \citep{chu2018meansum}.
Simple extractive algorithms developed upon graph-based ranking algorithms such as TextRank \citep{mihalcea-tarau-2004-textrank} can also be applied. 
However, if the articles in a group convey different stories, such methods built upon the assumption of overlapping information fail.
Graph-based ranking algorithms for extracting salient sentences also become suboptimal if directly applied since they can lose focus on the information, provided there are multiple at once.
In the case of summarizing a group of news articles, it is possible that some articles are conveying the same story. 
If the articles are summarized individually, the resulting group of summaries may still have overlapping and hence redundant information that should have been unified. 
If the articles are summarized based on the overlapping information, key information from other non-overlapping news articles is lost.
Hence, it is important to balance the need of preserving useful information, while casting away redundant, overlapping information.

This work explores the problem of summarizing a more generic group of documents, keeping the important information while retaining the least amount of redundancy, by adopting a two-stage framework. 
The first extractive stage is to collect salient information from each individual article, with the assumption that most articles have different stories from each other.
The second abstractive stage is to reduce redundant information from the previous extraction, where redundant sentences are merged to preserve the information while reducing the number of sentences, with the assumption that there exists a small number of articles with similar topics.
We demonstrate in \Cref{sec:experiments} that our approach with few exceptions outperforms all other baseline methods on evaluation datasets that match the aforementioned assumptions.
The third stage aims at replacing the repetitive entities such as names, locations, etc. incurred by the coreference resolution methods applied in the first stage with proper pronouns. 
Even though this stage does not affect the informativeness of the summary, it makes the summary sentences less redundant and more readable. 

The contributions of this paper are four-fold.
First, we, to the best of our knowledge, pioneer the exploration of abstractive summarization of multiple heterogeneous documents by proposing a novel, flexible three-stage framework independent of labeled data for tackling this task.
Second, we automatically construct datasets characterized by diverse information and provide the construction algorithm for future studies in this field.
Third, we augment the sentence fusion algorithm based on a word graph and the k-shortest path algorithm \citep{filippova-2010-multi} by employing a unidirectional language model, GPT2 \citep{radford2019language}, to dynamically compute the cost value based on the sentence probability, improving the quality of generated summary sentences. 
Lastly, we propose reverse coreference resolution (RCR) designed to replace redundant entity names with their corresponding pronouns, making the sentences more concise and easy to read. 

\section{Related Work}
The task of text summarization has been widely studied and it can be categorized from three perspectives: (i) single vs. multi-document summarization tasks, (ii) supervised vs unsupervised summarization methods and (iii) abstractive vs. extractive summarization forms.
The framework we propose is designed to perform abstractive text summarization on multiple documents in an unsupervised manner. 
Many efforts have been spent on supervised text summarization, as has been recapped by \citet{10.1145/3419106}.
These learning-based methods are proved to be effective based on benchmarking experiments.
However, they require large training data, which may not be available under many circumstances.
Therefore, we intend to focus on methods that do not require labeled data.

Efforts on unsupervised methods can also be grouped into two categories: (i) single-document summarization and (ii) multi-document summarization.
In the field of unsupervised single-document summarization, \citet{zheng-lapata-2019-sentence} employ BERT to better capture sentence semantics for similarity computation and propose to build a directed graph where each vertex is a sentence and the directed arcs indicate whether a sentence precedes or succeeds another sentence.
They argue that two sentences sharing similar information may have an asymetric relationship in that one sentence may make sense on its own, the other sentence may not as it depends on its peer.
Therefore, they propose to utilize directed sentence centrality for single-document summarization, where the centrality of a sentence or vertex in the graph is computed as the weighted sum of the weights of all its adjacent incoming and outgoing edges, instead of summing the weights of all its adjacent undirected edges. 
Despite that this method is developed for single-document summarization, it can be generalized to a multi-document setting, as we show in our experiment, where it is used as a benchmark.
\citet{dohare-etal-2018-unsupervised} propose to generate abstractive summaries by first generating AMR graphs (abstract meaning representation) for corresponding input stories, extract summary graphs from the AMR graphs, and lastly create summaries from the summary graphs.

In the field of unsupervised multi-document summarization, \citet{chu2018meansum} propose an end-to-end, unsupervised, abstractive, neural summarization model.
Their model consists of two parts: (i) an LSTM-based auto-encoder that learns a representation of each input text and (ii) a summarization module that learns to generate summaries based on the representation encoded by the auto-encoder.
The content of all input texts is obtained as the mean of all last hidden and cell states of the LSTM network.
This information is then fed into the decoder of the auto-encoder to decode a summary.
Lastly, they evaluate average similarity of summary to all input texts by encoding the summary again with the encoder of the auto-encoder and computing the average cosine distance between the hidden states of each encoded input text and the hidden state of the encoded summary.
Their approach requires no labeled training data to summarize multiple documents. 
However, it is also based on an assumption that all input texts to be summarized contain similar information. 
This is often not the case in real world as often times documents to be summarized have different gist. 
In contrast to neural-based summarization methods, \citet{nayeem-etal-2018-abstractive} propose an unsupervised multi-document summarization system comprising of word graph-based \citep{filippova-2010-multi} sentence fusion and integer linear programming (ILP)-based sentence ranking.
They first apply hierarchical agglomerative clustering with complete linkage to all sentences in the documents, where the distance between two sentences are based on their continuous representations.
For each cluster of related sentences, they generate abstractive fused sentences by first constructing a word graph and then apply k-shortest path on it.
They not only merge similar sentences into a single one but also substitute such lexicals as verbs or nouns with ones having functional similarities, making the resulting sentence more abstractive.
Then, they employ the concept-based ILP framework to find the best subset of sentences by designing an ILP formulation so that they cover as many important concepts as possible while ensuring the summary size is within some constraint.
\citet{summpip} take a similar approach. 
However, instead of directly grouping and merging similar sentences, they first construct a sentence graph, where a node corresponds to a sentence and an edge corresponds to certain connections between two sentences.
In addition to sentence similarity based on their continuous representations, this proximity can be determined based on whether one sentence refers to an entity in its preceding sentence, one sentence contains entities of the same type from the preceding sentence, or if one sentence follows another based on discourse markers. 
After that, they apply graph clustering to identify communities of sentence nodes in the graph. 

Even though our framework shares sentence clustering and sentence compression with the two aforementioned approaches, our approach is distinct from theirs in the following ways. 
First, they apply hierarchical clustering to all sentences, whereas we apply the clustering only to per-document summary sentences.
Second, they adopt a 3-gram language model during the word graph construction phase to ensure the substituted lexicals are meaningful, whereas we adopt a uni-directional language model during the k-shortest path phase to ensure the next token added to the path not only contains useful information but also ensures that the path has high linguistic probability.
Lastly, we skip the ILP process used to select useful sentences after sentence fusion and instead narrow down the salience for each document before sentence fusion.
This is because we hypothesize that a real-world group of multiple documents have no intrinsic connection to each other and hence performing information extraction before sentence fusion could preserve more salient information.

\section{Background}
In this section, we describe the preliminary work that our methodology is based on.

\subsection{TextRank}
The first stage of our framework is extracting key sentences from each document. 
We adopt TextRank, \citep{mihalcea-tarau-2004-textrank}, an extractive summarization algorithm.
Given a set of all sentences in a document, it first build an undirected graph $G = (\mathcal{V}, \mathcal{E})$ where $\mathcal{V}$ is the set of vertices and $\mathcal{E}$ the set of undirected edges; each edge $e_{ij}$ connects two vertices $v_i$ and $v_j$ and is assigned a weight $w_{ij}$. 
A vertex $v_i$ corresponds to a sentence in the document and the weight $w_{ij}$ corresponds to the similarity score between sentence $v_i$ and $v_j$.

Once the graph is built, the following formula introduced by \citet{mihalcea-tarau-2004-textrank} is applied to compute the score for each vertex
\begin{equation*}
  S^{k+1}(v_i) = (1 - d) + d \sum_{v_j \in N(v_i)} \frac{w_{ji}}{\sum_{v_k \in N(v_j)}}S^{k}(v_j),
\end{equation*}
where $S^{k}(v_i)$ denotes the score for $v_i$ at iteration $k$, $N(v_i)$ all neighbors of $v_i$, and $d$ the damping factor, which is set to $0.85$. 
The score for each vertex is set to $1.0$ at iteration $0$. 
Then, the score for each vertex is computed iteratively until convergence is reached. 
A convergence is achieved when the difference between the scores computed at two successive iterations for any vertex is smaller than a threshold $\tau_{pagerank}$.
Lastly, all vertices, i.e. sentences, are ranked by their scores and the top $r\%$ sentences are extracted as important information.

\subsection{Word Graph-based Sentence Fusion}
In the second stage, to merge a group of similar sentences into fewer ones, we adopt the word graph-based sentence fusion algorithm proposed by \citet{filippova-2010-multi}. 
Given a set of sentences $\mathcal{S} = \{s_1, ..., s_n\}$, the algorithm iteratively constructs a weighted, directed word graph $G = (\mathcal{V}, \mathcal{E})$, where each vertex $v_i \in \mathcal{V}$ corresponds to a word in a sentence along with its parts-of-speech (POS) tag and each edge $e_{ij} \in \mathcal{E}$ captures the word $v_i$ is followed by $v_j$ in a sentence. 
Initially, $G$ has two vertices: a dummy start and end vertex. 
At each iteration more vertices and edges are added to the graph from one sentence, where the start vertex is connected to the first word in the sentence and the last word in the sentence connected to the end vertex. 
If a word in a new sentence has the same POS tag as a word already existing in $G$, the vertex corresponding to this word is reused. 
The computation for the weight of each edge, detailed in \citep{filippova-2010-multi}, is designed to generate (i) grammatical paths based on strong links, i.e. it is significantly frequent for words to appear together, and (ii) informative path, i.e. paths containing salient words are encouraged.
After constructing the word graph, the algorithm employs the k-shortest path algorithm to find the top $k$ paths in the graph.

\subsection{Pointer Generator Network}
For the RCR stage, we designed a model based on the concept of the pointer generator network proposed by  \citet{see2017point}.
The goal of this network is to replace repetitive entity names with the proper pronouns, and retain the rest of the content by copying every other word. 
Let $\mathbf{h}^*_t$ be the weighted sum of encoder hidden states with attentions as weights, known as the context vector at time step $t$, $\mathbf{s}_t$ the decoder state at time step $t$, and $\mathbf{V}$, $\mathbf{V}'$, $\mathbf{b}$, $\mathbf{b}'$ learnable parameters.
A probability distribution over all words in the vocabulary $\mathcal{V}$ as possible output tokens is defined as
\begin{equation}
    P_{vocab} = \mathrm{softmax}(\mathbf{V}'(\mathbf{V}[\mathbf{s}_t, \mathbf{h}^*_t] + \mathbf{b}) + \mathbf{b}'),
\end{equation}
with $P_{vocab}(w)$ as the probability of word $w$.
For words $w$ out-of-vocabulary (OOV) $P_{vocab}(w)$ is set to $0$.
Following \citet{see2017point}, the generation probability $p_{gen}$ for the time step $t$ is defined as
\begin{equation}
    p_{gen} = \sigma (\mathbf{w}^\top_{h}\mathbf{h}^*_t + \mathbf{w}^\top_s \mathbf{s}_t + \mathbf{w}^\top_x\mathbf{x}_t + b_{ptr}).
\end{equation}
Lastly, the probability of a word that encompasses out-of-vocabulary (OOV) words is defined as
\begin{equation}
    P(w) = p_{gen}P_{vocab}(w) + (1 - p_{gen})\sum_{i:w_i=w}\mathbf{a}^t_i,
\end{equation}
where the attention distribution $a^t_i$ for the $i$-th time step is calculated as in \citet{bahdanau2014neural}:
\begin{equation}
    \mathbf{e}^t_i = \mathbf{v}^{\top} \tanh(\mathbf{W}_h \mathbf{h}_i + \mathbf{W}_s \mathbf{s}_t + \mathbf{b}_{attn}),
\end{equation}
\begin{equation}
    \mathbf{a}^t = \mathrm{softmax}(\mathbf{e}^t).
\end{equation}

\section{Methodology}
\begin{figure*}
    \includegraphics[scale=0.5]{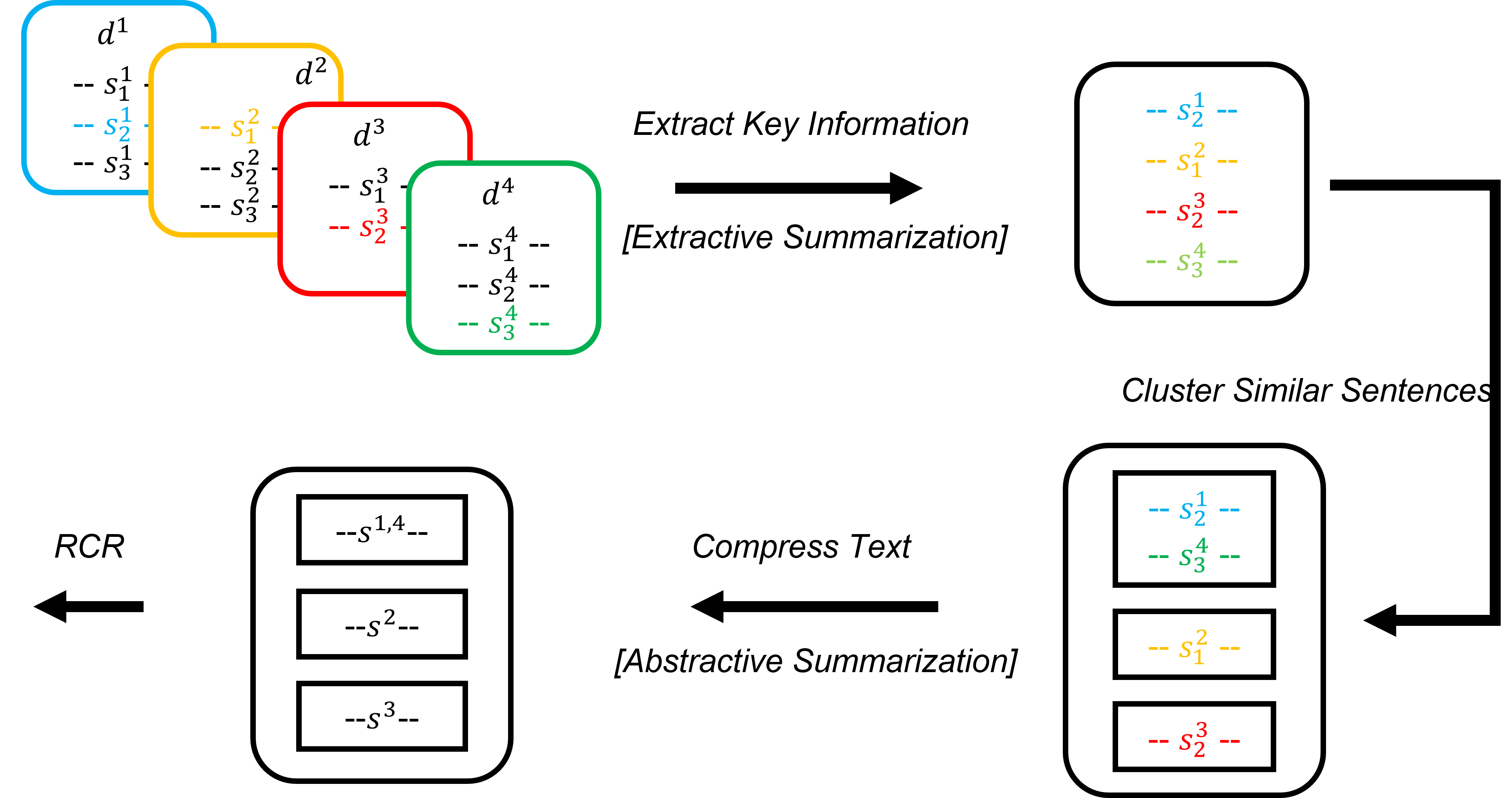}
    \centering
    \caption{
    An overview of the framework. 
    In the first stage, an unsupervised single-document summarization method, e.g. TextRank, is employed to extract useful information (i.e. sentences with non-black color) from each document.
    In the second stage, sentences with overlapping information (i.e. sentences with the same color) are first grouped with a clustering algorithm, e.g. a hierarchical agglomerative clustering algorithm.
    In the same stage, the clustered sentences are compressed with sentence fusion techniques, such as word graph and the k-shortest path-based algorithm, to reduce redundancy. The last stage is RCR.
    }
    \label{fig:overview}
\end{figure*}

In this section we describe the framework proposed for summarizing multiple documents that can contain different gist.
It consists of three stages, each with its own objective: (i) per-document information extraction, (ii) information redundancy minimization and (iii) RCR.
The first stage aims at identifying and extracting salient information from each coreference-resolved individual document. 
This is under the assumption that most documents in a document group, e.g. a group of news articles from a time period, are independent of each other. 
Therefore, the task of finding key information in one document is unrelated to finding key information in another. 
The second stage aims at identifying sentences with similar information and merging those sentences into one so that the final summary is free of redundancy. 
This is under the assumption that some documents in a document group discuss the same story and hence the extracted information from the first stage might be overlapping.
Lastly, the third stage aims at removing excessive mentions of entities due to the application of passage-wise coreference resolution. 
An overview of our framework is shown in \Cref{fig:overview}.

\subsection{Per-Document Information Extraction}
Before we extract sentences from a document, we need to make sure the extracted sentences do not contain ambiguous terms such as pronouns.
In order to replace such terms with meaningful entities, we adopt the coreference resolution model described in \citet{clark2016deep}.
After that, we employ the TextRank algorithm \citep{mihalcea-tarau-2004-textrank} with one modification to perform sentence extraction.
In the original implementation, the similarity score is calculated based on overlapping word.
In our implementation, we adopt BERT to embed the sentences, and then use the cosine similarity value computed from the sentence embedding vectors as the similarity score.
This is because BERT can capture more nuanced correlation between words and hence can map sentences with similar meanings but few overlapping words to close locations in the vector space.

\subsection{Information Redundancy Minimization}
There are two steps taken to remove redundant information: (i) sentence clustering and (ii) sentence merging.
The sentence clustering step aims at grouping similar sentences. 
We employ the hierarchical agglomerative clustering algorithm, similar to what is done by \citet{nayeem-etal-2018-abstractive}. 
One difference is that, instead of using the complete linkage criteria, we adopt the ward linkage criteria. 
Unlike the complete linkage criteria, which determines the proximity between two clusters based on the largest distance between a sentence in one cluster and a sentence in the other, the ward linkage criteria computes the proximity based on the magnitude by which the summed square in their joint cluster would be greater than the combined summed square in these two original clusters. 
This ensures that two clusters containing similar sentences are still merged even though there exists a small number of outliers. 
Furthermore, we adopt a multi-stage clustering procedure, where we start with a relatively large initial distance threshold to obtain $n$ clusters from all $m$ sentences.
If a cluster has more than $m/n$ sentences, we further fine-grain this cluster with a distance threshold half of the previous one, until the distance threshold reaches $0.5$.
The initial threshold value is denoted as $\tau_{cluster}$.

Whereas the common practice for the k-shortest path algorithm is to simply adopt the edge weights as costs, we add a dynamic costs based on sentence probability computed during the k-shortest path algorithm in order to further improve the grammacality of the generated sentences. 
Specifically, we adopt GPT-2 \citep{radford2019language}, a uni-directional language model, to compute a linguistic score based on each of the neighboring vertices of the last vertex of the current path.
The cost of adding a vertex $v$ to an existing path $p = [v_1, v_2, ..., v_{k-1}]$ becomes 
\begin{equation*}
    a \cdot w + (1-a) \cdot \frac{k}{\mathrm{log}P(v_k\, |\, [v_1:v_{k-1}])},
\end{equation*}
where $w$ is the edge weight proposed in \citet{filippova-2010-multi}, $P(v_k\, |\, [v_1:v_{k-1}])$ is the probability of $v_k$ given all previous words, and $a$ is a weight for combining these two pieces of information.
Notice that we invert the average log probability because the higher the probability of a sentence, the lower the cost should be for k-shortest path.

Furthermore, because a sentence cluster may still contain a few outlier sentences with other meanings, we design an algorithm that ranks all $k$ sentences based on their sentence probabilities and then keep all distinct sentences determined by the Ratcliff/Obershelp Pattern Recognition algorithm \citep{black2004ratcliff}. 
Two sentences are considered distinct from each other if their similarity score is smaller than $d$. 

\subsection{Reverse Coreference Resolution}
Replacing pronouns with actual entity names preserves key information from the original documents.
However, when a sentence contains repetitive entity names, it becomes redundant and less readable. 
Therefore, we propose a pointer generator network-based Reverse Coreference Resolution model (RCR) that takes a sentence with repetitive entities as input, and produces an output sentence with the extra entities replaced with proper pronouns. 

Our reverse coreference model consists of two parts, an encoder and an attention-guided decoder.
Given an input sentence containing $L$ tokens, the encoder embeds every token and produces a vector of encoder states. The hidden layer size is denoted as $hs$.
The decoder takes the embedding of the previously predicted token $\mathbf{x}_t$, previous hidden state $\mathbf{s}_t$ and all encoder states as input and computes the context vector $\mathbf{h}^*$ by applying weighted average of the encoder states, with the attention vector as the weights. 

Similar to the generation probability used in the pointer generator network, the decoder computes the replacement probability 
\begin{equation}
	p_{r} = \sigma(\mathbf{w}_h^{\top} \mathbf{h}^* + \mathbf{w}_s^{\top} \mathbf{s}_t + \mathbf{w}_x^{\top} \mathbf{x}_t),
\end{equation}
which determines whether or not at time step $t$ a word should be copied from the original text, or should be replaced with a word from a predefined set of $N$ replacement words such as pronouns, etc.
The replacement word distribution is
\begin{equation}
	\mathbf{o}_{r} = \mathrm{softmax}(\mathbf{W}_{r} (\mathbf{W}_{rh} \mathbf{h}^* + \mathbf{W}_{rs} \mathbf{s}_t + \mathbf{W}_{rx} \mathbf{x}_t)),
\end{equation}
where $\mathbf{W}_{rh}$, $\mathbf{W}_{rs}$, $\mathbf{W}_{rx} \in R^{hs \times hs}$, and $\mathbf{W}_{r} \in R^{hs \times N}$ are learnable weights. 
Similarly, the copy word distribution is 
\begin{equation}
	\mathbf{o}_{c} = \mathrm{softmax}(\mathbf{W}_{c} (\mathbf{W}_{ch} \mathbf{h}^* + \mathbf{W}_{cs} \mathbf{s}_t + \mathbf{W}_{cx} \mathbf{x}_t)),
\end{equation}
where $\mathbf{W}_{ch}$, $\mathbf{W}_{cs}$, $\mathbf{W}_{cx} \in R^{hs \times hs}$,
and $\mathbf{W}_{c} \in R^{hs \times L}$ are learnable weights.
The final distribution is defined as
\begin{equation}
	\mathrm{concat}([(1-p_{r}) \mathbf{o_{c}}; p_{r}\mathbf{o_{r}}]).
\end{equation}

The model is hence trained to identify if replacement should be done for a certain word, and if so, which word from the $N$ replacement words to choose from, and if not, what the index of the word in the original sentence that should be copied is. 
To obtain the training data for RCR, it suffices to use the output of the coreferencing model as input and use the original input sentences of the coreferencing model as ground truth.

\section{Experiments}
\label{sec:experiments}
 In this section, we present our experimental details for evaluating the performance of our framework compared with several other baseline approaches. We also discuss the decision behind designing a novel dataset for evaluating multi-document summarization and point out the disadvantages of some existing datasets.

\subsection{Dataset Generation}
Existing datasets on training or evaluating multi-document summarization systems such as MultiNews, Opinosis and DUC2004 have one major flaw -- each document group contains texts about a shared topic. 
Whereas they still provide insights into the performance of many multi-document summarization systems, they fail to assess the case where a document group contains various stories. 
In order to evaluate the efficiency of our system at summarizing multiple diverse documents, we propose two variations of existing multi-document summarization datasets that are built on top of existing datasets: the CNN/Daily Mail and Newsroom datasets.

The CNN/Daily Mail dataset \citep{nallapati2016abstractive} contains 311,672 online news article-summary pairs, where on average each article has 766 words, and each summary includes 53 words.
The Newsroom dataset \citep{N18-1065} has in total 1,321,995 article-summary pairs, where on average each article contains 658.6 words and each summary is composed of 26.7 words.

The generation process is described as follows.
Given a collection of documents $\mathcal{D}$ with $n$ documents, we triple the number of documents and obtain a new collection $\hat{\mathcal{D}}$.
The reason why we triple the amount of original documents is that we want to mimic real world cases where a large batch of documents may contain a few documents discussing similar topics.
In our case, documents discussing similar topics are the potentially duplicated documents. 
Then, we generate $m$ groups of documents by randomly drawing $s$ documents from $\hat{\mathcal{D}}$ at a time.
Here $s$ denotes the scale, i.e. how many documents each group has.
Lastly, for each group, we generate the summary by concatenating the 
summary of each corresponding document. 
However, even though there can be duplicated documents in a group, we only keep one copy of the summary of such documents.
This step is taken to ensure that summarization systems being evaluated can identify redundant information and avoid including such information twice in the summary.
For each variation, CNN/Daily Mail and Newsroom, we randomly generate three batches of datasets with three different random seeds, and each batch has two different scales: $s$ of $10$ and $100$.
Therefore, there are in total $2 \cdot 3 \cdot 2 = 12$ datasets.
In addition, we use $m=100$ for $s=10$ and $m=10$ for $s=100$.
We summarize in \Cref{tab:data-stats} the statistics of the generated datasets.

\begin{table*}[t]
\centering
\caption{A brief statistics of the dataset. The unit of length is the number of words. Notice that the length of the articles and summaries are proportional to the scale of the dataset. }
\begin{tabular}{@{}lrrr@{}}
\toprule
Dataset                  & Mean Article Length & Mean Summary Length &  \\ \midrule
CNN/Daily Mail Scale 10  & 7575.49             & 494.70              &  \\
CNN/Daily Mail Scale 100 & 75754.87            & 4946.40             &  \\
Newsroom Scale 10        & 7411.31             & 301.35              &  \\
Newsroom Scale 100       & 74113.11            & 3013.46             &  \\ \bottomrule
\end{tabular}

\label{tab:data-stats}
\end{table*}

\subsection{Baseline systems and Results}


\begin{table*}[]
\centering
\caption{Benchmarking results. Each group of three numbers correspond to the Rouge-1, Rouge-2 and Rouge-L scores. Our models consistently outperform other unsupervised methods except for Rouge-L scores for the CNN dataset, where there are ties with AUMDSPSF.}
\begin{tabular}{@{}lrrrr@{}}
\toprule
Summarization Models/\\Algorithms & \multicolumn{1}{l}{CNN-10} & \multicolumn{1}{l}{CNN-100} & \multicolumn{1}{l}{NewsRoom-10} & \multicolumn{1}{l}{NewsRoom-100} \\ \midrule
LSMD                            & \textbf{0.30 0.07 0.27}    & \textbf{0.47 0.11 0.38}     & \textbf{0.24 0.09 0.21}         & \textbf{0.33 0.12 0.29}          \\
LSMD+RCR                        & \textbf{0.30 0.07 0.27}    & \textbf{0.47 0.11 0.38}     & \textbf{0.24 0.09 0.21}         & \textbf{0.33 0.12 0.29}          \\
AUMDSPSF                        & 0.27 \textbf{0.07 0.27}    & 0.38 \textbf{0.11 0.38}     & 0.16 0.08 0.19                  & 0.19 0.08 0.23                   \\
PacSum                          & 0.23 0.07 0.25    & 0.27 0.09 0.34              & 0.16 0.09 0.16                  & 0.18 0.08 0.23                   \\
CBCWE                           & 0.24 0.06 0.23             & 0.33 0.09 0.35              & 0.17 0.06 0.16                  & 0.21 0.08 0.23                   \\
TextRank                        & 0.23 0.06 0.24             & 0.29 0.09 0.36              & 0.16 0.06 0.17                  & 0.19 0.08 0.23                   \\ \bottomrule
\end{tabular}

\label{tab:results}
\end{table*}

\begin{table*}[t]
\centering
\caption{Some example no-RCR and simplified sentences produced by a trained reverse coreference model. }
\begin{tabular}{@{}lll@{}}
	\toprule
	Index & No-RCR Sentences                                                                                                                                                       & Simplified Sentences                                                                                                                                                     \\ \midrule
	1     & \begin{tabular}[c]{@{}l@{}}trump responded using one of trump \\ favorite platforms , logging onto twitter \\ in the early morning hours \end{tabular}                   & \begin{tabular}[c]{@{}l@{}}trump responded using one of his \\ favorite platforms , logging onto twitter \\ in the early morning hours \end{tabular}                    \\ \midrule
	2     & \begin{tabular}[c]{@{}l@{}}watch a former michael jackson's \\ guitarist reflect on michael jackson \\ 's career\end{tabular}                                             & \begin{tabular}[c]{@{}l@{}}watch a former Michael Jackson 's \\ guitarist reflect on his career\end{tabular}                                                             \\ \midrule
	3     & \begin{tabular}[c]{@{}l@{}}rather than helping out those in the \\ toughest shape , it looks like democrats \\ ended up helping democrats supporters\end{tabular}       & \begin{tabular}[c]{@{}l@{}}rather than helping out those in the \\ toughest shape , it looks like democrats \\ ended up helping their supporters\end{tabular}            \\ \midrule
	4     & \begin{tabular}[c]{@{}l@{}}george was arrested in Texas this \\ weekend after cops say george assulted \\ george wife \end{tabular}                                      & \begin{tabular}[c]{@{}l@{}}george was arrested in Texas this \\ weekend after cops say he assulted his \\ wife \end{tabular}                                            \\ \midrule
	5     & \begin{tabular}[c]{@{}l@{}}two scientists relied on two distinct \\ features of ancient settlements in the \\ near east : soils altered by human \\ activity\end{tabular} & \begin{tabular}[c]{@{}l@{}}these scientists relied on two distinct\\ features of ancient settlements in the\\ near east : soils altered by human\\ activity\end{tabular} \\ \midrule
	6     & \begin{tabular}[c]{@{}l@{}}but apparently, kim achieved kim tiny \\ waist with one of the Kardashians' \\ beloved korsets\end{tabular}                                    & \begin{tabular}[c]{@{}l@{}}but apparently , kim achieved his tiny \\ waist with one of the kardashians ' \\ beloeved korsets \end{tabular}                              \\ \bottomrule
\end{tabular}
\label{tab:reverse-coref-examples}
\end{table*}


We evaluate our summarization framework using the ROUGE metrics. 
Particularly, we report Rouge-1, Rouge-2 and Rouge-L for each dataset and algorithm.
We evaluate our framework, with (LSMD+RCR) and without RCR (LSMD) against all state-of-the-art unsupervised systems such as AUMDSPSF proposed by \citet{nayeem-etal-2018-abstractive}, CBCWE proposed by \citet{rossiello-etal-2017-centroid}, PacSum proposed by \citet{zheng-lapata-2019-sentence} and TextRank proposed by \citet{mihalcea-tarau-2004-textrank}.

Whereas most systems have disclosed their source codes, we have to implement the AUMDSPSF ourselves.
As described by \citet{nayeem-etal-2018-abstractive}, we first group all sentences in the documents using the hierarchical agglomerative clustering algorithm with complete linkage and the distance threshold of $0.5$. 
The sentence embedding is obtained through the encoder of a bi-directional GRU-based auto-encoder, trained on the English monolingual news dataset obtained from statmt.org.\footnote{http://www.statmt.org/wmt13/translation-task.html}
Then, we construct a word graph for each sentence cluster with lexical substitution, run k-shortest path to get the top 50 sentences, rank those sentences with TextRank and assign each sentence a rank score.
Lastly, we compute the score for each candidate according to \citet{nayeem-etal-2018-abstractive} and select the top abstractive sentences based on their ILP formulation.

We follow \citet{mihalcea-tarau-2004-textrank} and set $\tau_{pagerank}$ to be $0.85$.
However, instead of taking the top r\% of sentences as the most informational content of an article, we retain top sentences with rank scores higher than that of the immediately next sentence, which is typically significantly lower.
Then, we set the initial threshold value for clustering to $\tau_{clustering}=2.0$ as empirical studies show that this results in the best balance of running time and performance. 
Furthermore, we set $k$ to be 10, because each sentence group under our framework has fewer sentences than the sentence groups do in \citet{nayeem-etal-2018-abstractive}.
Lastly, we set the distance threshold $d$ used to determine text similarity to $0.3$. 
With this setup, our algorithm produces summaries with 47.6\% abstractive sentences.
The rest of the summaries remain unmodified regarding the source texts, as they contain distinct information that is not grouped into clusters and then merged. 

We run experiments on all batches of datasets with all methods.
Since each Dataset-Scale pair has three corresponding variations generated with three different random seeds, we summarize the result by taking the average of all Rouge-1, Rouge-2, and Rouge-L scores for each Dataset-Scale pair.
We present in \Cref{tab:results} the benchmarking results. 
Our framework consistently outperforms all other unsupervised methods. 
The improvement is more significant when the dataset scale is larger, i.e. when there are more heterogeneous documents to be summarized at once. 
Specifically, for the CNN-based datasets, our model outperforms the second best system (AUMDSPSF) by $15\%$ for scale of 10 and by $24\%$ for scale of 100 in terms of the Rouge-1 scores.
Similarly, for the NewsRoom-based datasets, our model outperforms the second best system by $59\%$ for scale of 10 while by $62\%$ for scale of 100 in terms of the Rouge-1 scores.
This indicates that our method is able to preserve key information in terms of unigram tokens regardless of their orders, while also ensuring limited redundancy.
In terms of the Rouge-L scores, our method does not significantly outperform the existing systems. 
This indicates that, when it comes to in-sequence co-occurrences of longest common subsequences, our model does not necessarily generate summaries that overlap significantly more with the ground truth summaries. 

\begin{figure}
    \includegraphics[scale=0.4]{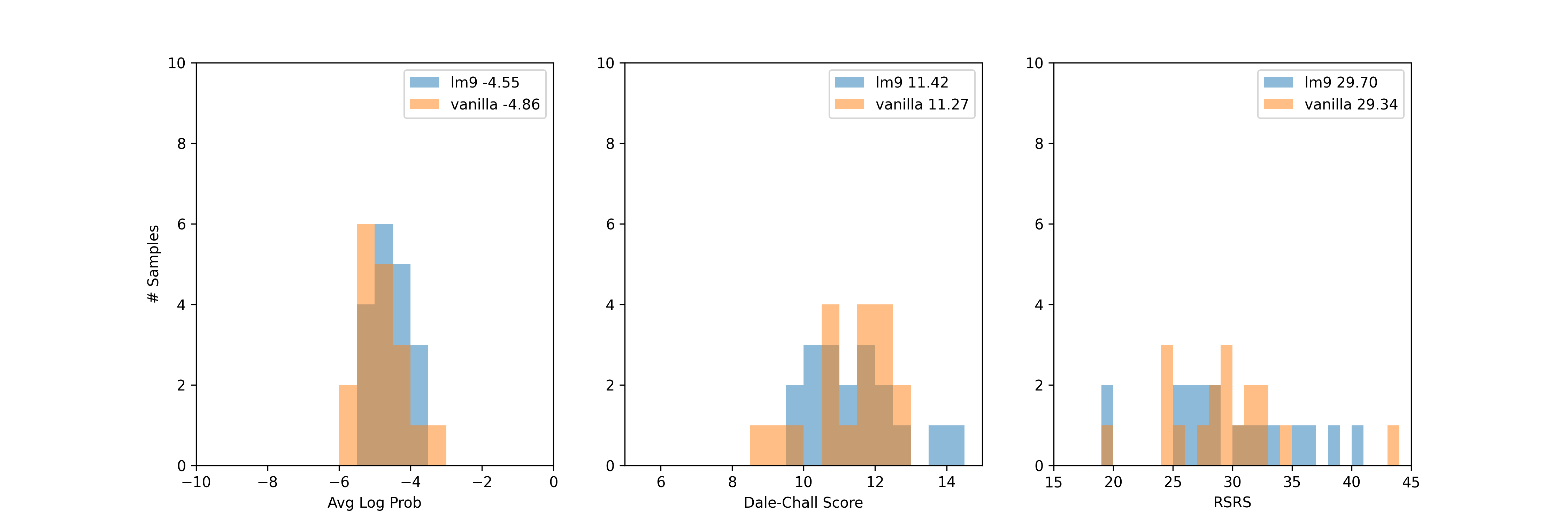}
    \centering
    \caption{
    Comparison among the average sentence log probabilities (the higher the better) and two variances of readability scores (the lower the better) of summaries of the newsroom-seed100 dataset generated with the language model-based k-shortest path and the vanilla k-shortest path.
    }
    \label{fig:readability}
\end{figure}

Furthermore, we compare two different readability scores, the Dale-Chall readability score \citep{dale1948formula} and the Ranked Sentence Readability Score \citep{martinc2021supervised}, as well as the average sentence probability computed on our summaries.
We notice in \Cref{fig:readability} that both the readability scores and average probabilities of summaries with sentences merged by the language model-enhanced k-shortest path algorithm are slightly higher than those by the vanilla k-shortest path algorithm.
This may seem unusual at first, as to one's instinct more complex sentences (i.e. those with \textit{higher} readability score) should correspond to higher average sentence probability.
However, one should also realize that both readability formulas utilized take the sentence length into account; specifically, longer sentences may lead to higher scores.
It is likely that even though a sentence or a summary has a higher readability score, it does not necessarily mean that it is harder to comprehend.
After all, readability scores are used mainly to classify texts for people of different educational level, and for adult readers, minor difference in readability score should not be noticeable.
On the other hand, average sentence probability produced by a well-trained GPT-2 reflects the probability of a word occurring after a sequence of words. 
A sentence with a higher probability means it is more likely similar sentences or phrases exist in the training dataset, indicating the sentence may be more grammatical.
Therefore, we conclude that our language model-enhanced k-shortest path algorithm helps promotes grammatical merged sentences, albeit it produces summaries with similar Rouge scores to the ones by the vanilla k-shortest path algorithm.

\subsection{Reverse Coreference Resolution Examples and Discussion}
We train and test our pointer generator network-based model on sentences with up to 20 tokens, including punctuation. 
While our model is able to generate some readable sentences that match the semantics of the original redundant correspondants, only $37.6\%$ of the validation examples result in exact word-for-word match with our ground truth sentences. 
That said, summaries processed with the RCR model produce better Rouge scores during evaluation, even though the improvement is tiny (i.e. to the third or fourth significant digit). 

The main drawback of our approach is poor handling of variable-length input sequence. 
Even though we use padding to make sure all input sequences are of the same length, i.e. 20 tokens, at decoding time, instead of predicting the padding token \texttt{PAD}, our model copies words from indices it has already copied from.
This results in sentences with repeating words.
We alleviate this problem by limiting the length of the output sequence to be the length of the input sequence.
This reduces the likelihood of having repeating words, but does not completely prevent them from occurring.
Furthermore, this problem worsens when we try training the model on longer sequences.
This indicates that our current model is not able to process long redundant sentences well.

We present in \Cref{tab:reverse-coref-examples} some example simplified sentences by our reverse coreference model. 
The first four examples showcase the ideal simplification performed by the model. 
It is able to learn the correct pronoun for people and even for political parties. 
The last two examples reveal some problems.
In example 5, the model mistakenly replaces ``two'' with ``these,'' which is not only unnecessary but it also alters the original semantics of the sentence. 
This may be due to noisy training data which are generated automatically using the existing coreference resolution model. 
In example 6, our model fails to assign the correct pronoun to replace ``kim'' in the original sentence.
In this particular case, the pronoun ``she'' should be used in place of ``he.''
We believe this is due to the name ``kim'' not only being used by male but also female human beings. 
Having seen this name occur most of the time with the masculine pronoun ``he'' or ``him,'' the model learns biased information about the gender of the name. 

\subsection{Applications on Real World Data}
We apply our method to news articles grouped by weeks as well as medical articles related to the corona virus.
We describe here the statistics of the datasets and the generated summaries.

\paragraph{Corona Virus Medical Research Paper Query}
The coronavirus has severely endangered public health and damaged economies.
To help fight the corona virus, with our summarization approach, we build a query system named CAVIDOQS (CoronA Virus DOcument Query System) for corona virus related research papers in the CORD-19 dataset \citep{wang2020cord19}.
In short, users can type in medicine-related entities as key words, and the system proposes a list of summary sentences containing those keywords, where each summary sentence can be traced back to articles comprising the corresponding information.
Because each summary sentence comprises the key information of the corresponding articles, we believe this enables researchers to quickly locate the paper that is most relevant to their work.
At the time we built CAVIDOQS, the dataset had 29,319 articles, even though it has since grown.
Our approach is ideal for this dataset, because, unlike other methods, ours does not assume that the documents to be summarized are discussing the same event or topic. 
For example, an article in the CORD-19 dataset can be discussing the effect of a particular protein on damping the replication of the SARS virus while another article can be discussing the experiment of a vaccine on a group of people whose ages are from 30 to 50. 
Because different medical researchers have different focus, directly applying existing multi-document summarization approaches to CORD-19 may not result in an acceptable outcome, as it is hard for a system to automatically determine which content is more important for one researcher and which content for another.

We first apply our summarization approach to bring down the number of sentences from over 300,000 to 55,721 summary sentences. 
This still leaves us with a large number of sentences to consider. 
Therefore, on top of the summaries, we group them by certain keys. 
In the case of the CORD-19 data, we identify various named entities using a named entity recognition model trained on medical texts provided by ScispaCy \citep{scispacy}.
Each key relates to summary sentences containing it, and works as the base query term of our system.

\paragraph{News Articles Grouped by Week} 
The aim of summarizing news articles grouped by weeks is to provide financial analysts a brief overview of recent events in the past weeks. 
Indeed, this research has been initiated by a large investment firm which also provided this dataset. 
We obtained news articles in three categories: communication, finance and tech. 
Each category has 102 weeks of articles, where each week relates to on average 589 articles for communication, 625 for finance and 668 for tech. 
There are on average 181,578 words and 7,091 sentences in a week of articles for the category communication, 189,430 words and 7,513 sentences for the category finance, and 177,351 words and 7,109 sentences for the category tech. 
Our unsupervised multi-document summarization approach manage to compress the information down to on average 6283 words and 151 sentences for a week of articles for the category communication, 6982 words and 167 sentences for the category finance, and 6358 words and 153 sentences for the category tech.
It dramatically reduces the amount of information and let a reader grasp the key information more efficiently.

\section{Conclusion and Future Work}
In this work, we first discuss the current limitations of existing multi-document summarization methods as well as datasets.
We proceed to propose a flexible three-stage framework to solve the more generic task of summarizing documents with heterogenous content, which is proved by our experiments to be better than existing unsupervised methods.
For the first information extraction stage, we employ the TextRank algorithm.
For the second information condensation stage, we employ hierarchical clustering to group similar information together, and the word graph and k-shortest path-based sentence fusion algorithm to merge similar information.
Lastly, for the third repetition removal stage, we propose a pointer generator network-based reverse coreference model that, instead of replacing pronouns with the corresponding entities to reduce ambiguities, replaces repetitive entities with appropriate pronouns to make the sentences more concise and readable.
To evaluate our framework in a more generic setting, we propose a simple data generation algorithm that generates multi-document datasets based on existing single-document summarization datasets.

An interesting future work can be devising a supervised end-to-end model comprising of all three stages by using a graph neural network to condense sentences instead of the traditional k-shortest path algorithm. 

\section{Acknowledgement}
We would like to express our gratitude toward Principal Financial for inspiring us with the concept of multi-document summarization with the underlying use case, and providing a dataset to experiment with our idea.
We are obliged to Dr. Joseph Byrum, Chief Data Scientist at Principal Financial Group for initiating this research. 
We are also indebted to Zack Amato and Talha Naushad for frequent guidance and subject mattered discussions.

%

\newpage
\clearpage

\bibliography{allbib}
\bibliographystyle{acl_natbib}

\end{document}